\documentclass[letterpaper, 10 pt, conference]{ieeeconf}
\IEEEoverridecommandlockouts
\usepackage{amsmath,amssymb,amsfonts}
\usepackage{algorithm}
\usepackage{algpseudocode}
\usepackage{graphicx}
\usepackage{textcomp}
\usepackage{xcolor}
\usepackage{units}
\usepackage{booktabs}
\usepackage{multirow}
\usepackage{subfig}
\usepackage{stfloats}
\usepackage{cite}
\usepackage{makecell}

\title{\LARGE \bf
Improving Driver Satisfaction with a Driving Function Learning\\from Implicit Human Feedback - a Test Group Study
}

\author{Robin Schwager$^{*,1}$, Andrea Anastasio$^{*,2}$, Simon Hartmann$^{*,3}$, Andreas Ronellenfitsch$^{1}$,\\Michael Grimm$^{1}$, Tim Brühl$^{1}$, Tin Stribor Sohn$^{1}$, Tim Dieter Eberhardt$^{1}$, and Sören Hohmann$^{4}$% <-this % stops a space
\thanks{$^{*}$Authors contributed equally.}
\thanks{$^{1}$Dr. Ing. h.c. F. Porsche AG, Weissach, Germany.
        {\tt\small (robin.schwager1, andreas.ronellenfitsch, michael.grimm3, tim.bruehl, tin\_stribor.sohn, tim.eberhardt1) @porsche.de}}%
        \thanks{$^{2}$University of Catania, Catania, Italy.
        {\tt\small anastasio.andrea@studium.unict.it}}%
        \thanks{$^{3}$Technical University of Munich, Munich, Germany.
        {\tt\small simon.hartmann@tum.de}}%
        \thanks{$^{4}$Karlsruhe Institute of Technology, Karlsruhe, Germany.
        {\tt\small soeren.hohmann@kit.edu}}%
}

\begin{document}

\maketitle

\bibliographystyle{IEEEtran}

\thispagestyle{empty}
\pagestyle{empty}

\begin{abstract}

During the use of advanced driver assistance systems, drivers frequently intervene into the active driving function and adjust the system's behavior to their personal wishes.
These active driver-initiated takeovers contain feedback about deviations in the driving function's behavior from the drivers' personal preferences.
This feedback should be utilized to optimize and personalize the driving function's behavior.

In this work, the adjustment of the speed profile of a Predictive Longitudinal Driving Function (PLDF) on a pre-defined route is highlighted.
An algorithm is introduced which iteratively adjusts the PLDF's speed profile by taking into account both the original speed profile of the PLDF and the driver demonstration.
This approach allows for personalization in a traded control scenario during active use of the PLDF.

The applicability of the proposed algorithm is tested in a driving simulator-based test group study with 43 participants.
The study finds a significant increase in driver satisfaction and a significant reduction in the intervention frequency when using the proposed adaptive PLDF.
Additionally, feedback by the participants was gathered to identify further optimization potentials of the proposed system.

\end{abstract}

\section{Introduction}\label{sec:introduction}

Automated and assisted driving functions are commonly seen in state-of-the-art vehicles.
Most in-production driving functions are categorized as levels 1 and 2 of the Society of Automotive Engineers (SAE) standard \cite{SAE2021}.
Each driving function has a clearly defined Operational Design Domain (ODD) within which either the longitudinal control, lateral control, or both are handled by the system without the need for driver intervention.
Drivers are only required to observe the driving task and take over control when the driving functions leaves its ODD or to correct potential mistakes made by the function.
However, during real-world driving, driver-initiated takeovers of control are commonly seen even within the ODD \cite{Schwager2024_study, Gershon2021, Morando2020}.
These takeovers of vehicle control are optional and voluntarily conducted by the driver, in contrast to the mandatory takeovers defined by the function's ODD \cite{Lu2016}.

The driving function that is the focus of this work is an in-production level 1 Predictive Longitudinal Driving Function (PLDF).
The term \textit{driver intervention} will be used exchangingly for short-term driver-initiated takeovers of vehicle control.
Adjustments of the driving function's set speed are also counted as driver interventions.

Reference \cite{Schwager2024_study} conducted a test group study, using the same PLDF as in this paper, in which the reasons behind driver interventions were manually annotated by the participants during naturalistic driving.
The most common reasons behind interventions were voluntary adjustments of the system behavior in free-driving scenarios based on deviating personal preferences by the drivers.
These frequent takeovers within the ODD show an optimization potential where the PLDF could be adjusted to better suit the individual drivers' wishes.
Reference \cite{Schwager2024_needed_adjustments} also found that drivers consistently intervene in specific locations but not necessarily do so in other comparable locations.
Therefore, a location-based speed profile adjustment is proposed.

Building upon these findings, this work proposes an algorithm for the iterative adjustment of the PLDF's free-driving speed profile in specific locations based on recorded driver interventions.
The iterative algorithm developed for the derivation of new target speed profiles combines the original PLDF speed profile with recorded driver interventions, while compensating for delayed and overreactive driver behavior.
In doing so, the algorithm aims to derive the underlying driver intention of the intervention instead of directly copying the recorded behavior.
The algorithm is evaluated in a driving simulator-based test group study where the same route is driven multiple times, and the PLDF's speed profile is adjusted iteratively based on recorded driver interventions.
The goal of the test group study is to investigate how drivers react to such an adaptive driving function and whether it increases driver satisfaction.

\section{Related work}\label{sec:related_work}

\subsection{Driver-Initiated Takeovers}

Driver-initiated takeovers can be distinguished into \textit{mandatory} and \textit{voluntary interventions}, depending on whether the driving function's design demands an intervention in a certain situation or whether a driver voluntarily decides to take over the vehicle control \cite{Lu2016}.
While most related studies focus on mandatory driver-initiated takeovers, this work focuses on voluntary interventions.

\subsubsection{Voluntary Interventions}

Drivers may voluntarily decide to take over the vehicle control within a driving function's ODD due to a variety of reasons \cite{Lu2016}.
In \cite{Schwager2024_study}, driver interventions during naturalistic driving were recorded in a test group study.
It was found that most driver-initiated takeovers were non-mandatory interventions to adjust the system behavior to the drivers' wishes during non-traffic-related situations.
This includes both adjustments to the PLDF's set speed, as well as gas and brake pedal presses to adjust the vehicle speed in scenarios such as curves, turns, and speed limit changes.
Furthermore, \cite{Schwager2025_classification} found that the mandatory and voluntary driver interventions in the dataset could be successfully automatically distinguished using neural network-based classifiers.

\subsubsection{Optimization Potential}

From the study results, \cite{Schwager2024_study} calculated a significant negative correlation between the frequency of these non-mandatory interventions and the drivers' satisfaction values stated in a questionnaire.
Therefore, they propose the optimization of driving functions based on the feedback contained in these non-mandatory interventions.
Further analyses of the recorded dataset showed that individual drivers intervened consistently only in specific locations, but not in all similar location types, e.g., only intervening consistently in a few specific curves instead of all curves on the drivers' commutes \cite{Schwager2024_needed_adjustments}.
Therefore, instead of changing the PLDF's general behavior, a location-based individualized adjustment is proposed, changing the drivers' speed profiles only in segments where they consistently intervene.

\subsection{Driving Function Personalization}

It is generally accepted in the state of the art that individual human driving styles can differ significantly \cite{Sagberg2015}.
Therefore, the personalization of Advanced Driver Assistance Systems (ADAS) should lead to higher driver satisfaction when the function behavior more closely follows the drivers' preferences \cite{Hasenjager2020}.

According to \cite{Hasenjager2020}, personalization can either be done explicitly by demanding specific feedback from the driver, or implicitly by observing driver behavior and deriving needed adjustments from that.
In the field of driving functions, implicit personalization is the prevalent approach.

\subsubsection{Driver Model Generation} \label{subsubsec:driver_model_generation}

Most commonly, driver models are generated from past recordings of manual driving.
These models have the goal to accurately imitate a driver's manual driving behavior. % \cite{Hasenjager2020, Wang2014, Kuderer2015, Kim2020, Bolduc2019, Wang2013, Lefevre2016, Chen2017}.
In the field of longitudinal driving functions, related work almost exclusively focuses on the personalization of Adaptive Cruise Control (ACC), with a focus on the derivation of parameters such as preferred headway, accelerations, jerks, and more \cite{Hasenjager2020, Wang2014, Kuderer2015, Wang2013, Lefevre2016, Chen2017}. % removed: Kim2020

\subsubsection{Limitations}\label{subsubsec:limitations}

These methods assume that drivers prefer their manual driving style also as the driving style of their ADAS.
However, state-of-the-art research is not conclusive whether this statement holds true \cite{Hasenjager2020}.
While \cite{Griesche2016,Ma2021} argue that drivers prefer a driving style which resembles their own, \cite{Basu2017,Yusof2016} found that all drivers prefer a more defensive ADAS driving style.

Another limitation of most driver models is their one-shot nature which relies on training on a fixed set of recorded demonstrations.
The goal of driving function personalization should be to increase the driver satisfaction while using a driving function.
However, related approaches fail to evaluate the effects of their personalization on the driver satisfaction.
Instead, they commonly evaluate how well their driver model copies the recorded manual driving style \cite{Hasenjager2020, Wang2014, Kuderer2015, Wang2013, Lefevre2016, Chen2017}.

References \cite{Hasenjager2020, Schwager2024_study, Schwager2024_needed_adjustments, Chen2017} therefore propose a continuous process of driving function personalization.
Instead of imitating manual driving styles, data for personalization should be collected during active ADAS use in real-world driving scenarios.
By interacting with the system, the driver provides feedback about their preferred driving style while the ADAS is active, which might differ from their manual driving style.
However, such a continuously updating personalized driving function based on driver's interactions with the active system could not be found in related work.

\subsection{Interactive Imitation Learning}
Interactive Imitation Learning (IIL) is an approach in the field of imitation learning which aims at learning from human feedback.
In IIL approaches, a human expert and the agent are in a traded control scenario where the agent executes its policy and the human expert may take over control of the agent to correct potential mistakes online.
After a test run with interventions by the human expert, the policy is updated accordingly \cite{Celemin2022}.

While this approach comes close to the goals of this paper, there are also multiple challenges associated with it.
For example, IIL approaches only perform well with consistent interventions by a human expert \cite{Celemin2022,Spencer2022}.
However, when used by an average human driver in a field setting, this consistent behavior cannot be expected \cite{Schwager2024_needed_adjustments}.
Also, the algorithms again focus on learning a general driving policy, similar to the driver models explained in chapter \ref{subsubsec:driver_model_generation}.
Therefore, IIL is not used in this work, but the methodology about the processing of human intervention behavior is still applicable.

\subsubsection{Description of Intervention States}\label{subsubsec:intervention_states}

Expert Intervention Learning (EIL), introduced in \cite{Spencer2022}, is an approach from the field of IIL.
In their work, the authors introduce a way to describe different parts of the driven trajectory with expert interventions.
First, there are the \textit{good states} where no intervention took place.
In these states, it is assumed that the agent performs well enough for the human operator to accept its behavior without the need to intervene.
These states are positively reinforced during the iterative training process.
When the agent leaves these good states by making a mistake, the human expert is required to take over control of the agent and correct said behavior.
When doing so, the human response is generally delayed by reaction times and the resulting intervention trajectory does not contain solely desirable states.
Thus, it is defined that the so-called \textit{bad states} begin shortly before an expert intervention starts and end shortly before the expert relinquishes control again.
These bad states are negatively reinforced during policy training.

A similar approach for the compensation of delayed expert interventions and undesirable recovery trajectories is introduced in \cite{Bi2020}.
There, an interpolation of the trajectory around the intervention is proposed to generate a smoother trajectory, which is then used to update the driving policy.

\subsection{Resulting Research Gap}

As explained in the sections above, most related work focuses on one-shot personalization approaches which aim at mimicking the manual driving style of drivers \cite{Hasenjager2020, Wang2014, Kuderer2015, Wang2013, Lefevre2016, Chen2017}.
However, it is not confirmed that drivers even prefer their manual driving style also as their automated driving style \cite{Griesche2016,Ma2021,Basu2017,Yusof2016}.
Therefore, it is proposed to adjust the driving function based on the driver feedback which is provided while actively using the driving function, similarly to how expert interventions are utilized in IIL approaches \cite{Celemin2022, Spencer2022}.
This approach builds upon related work that states the need for continuous driving function personalization approaches \cite{Hasenjager2020, Schwager2024_study, Schwager2024_needed_adjustments, Chen2017}, and related work that identified driver interventions as source for optimization potential \cite{Schwager2024_study,Schwager2024_needed_adjustments}.

Since the goal of personalization is to increase the driver's satisfaction, the personalized driving function should be tested and evaluated by the drivers after the personalization.
In related work, this step is generally skipped and the personalized model's performance is instead evaluated by how well the manual driving style is mimicked \cite{Hasenjager2020, Wang2014, Kuderer2015, Wang2013, Lefevre2016, Chen2017}.

Furthermore, related work generally focuses solely on the personalization of ACC driving \cite{Hasenjager2020, Wang2014, Kuderer2015, Wang2013, Lefevre2016, Chen2017}. %, and no research on the individualization of a driving function's behavior in free-driving scenarios could be found.
However, references \cite{Schwager2024_study,Schwager2024_needed_adjustments} found that drivers most commonly intervene during free driving scenarios in specific locations.

Based on these shortcomings, a personalized driving function is proposed in this paper, which aims to iteratively learn from active driver interventions during free driving scenarios by updating local speed profiles.
It is investigated whether utilizing driver interventions in this way may increase the driver satisfaction.

\section{Driving Function Speed Profile Adjustment} \label{sec:driving_function_speed_profile_adjustment_methodology}

This chapter details the methodology behind the individualized driving function.
Specifically, the original PLDF is explained in more detail and the adjustment of the speed profile based on driver interventions is explained.

\subsection{Predictive Longitudinal Driving Function}
The driving function used in this study is a PLDF that extends the functionality of traditional ACC.
Its implementation mirrors the behavior of an in-production SAE level 1 PLDF.
The PLDF takes into account map information about the upcoming route and calculates an appropriate speed profile, hence it is called ``predictive''.
It always abides by the legal speed limit by automatically setting detected speed limits as its set speed.
It also decelerates predictively onto lower legal speed limits to reach the speed limit sign at exactly the according speed.
Conversely, when approaching higher speed limits, the PLDF accelerates only after passing the higher speed limit.
In curves, turns, and roundabouts, the PLDF calculates an appropriate speed depending on the curvature of the road and decelerates accordingly.

\subsection{Driver Interventions} \label{subsec:driver_interventions}
While using the PLDF, the driver can intervene and take over control of the vehicle by pressing either the gas or the brake pedal.
When pressing the gas, the system behavior is overridden and the vehicle accelerates until the driver releases the pedal and the PLDF reactivates.
When pressing the brake, the vehicle decelerates and the PLDF deactivates until the driver activates it manually again.
Additionally, drivers can adjust the speed driven on straight segments by increasing or decreasing the set speed, e.g., by adding an offset of plus or minus \unit[5]{km/h}.
This offset is then discarded when reaching the next higher or lower speed limit.
It is important to note, however, that the base PLDF does not record these interventions, i.e., the system must be adjusted again each time a specific road is driven.

\subsection{Speed Profile Adjustment Algorithm}

To enable the PLDF to learn the drivers' individual preferences on a pre-defined route, a Speed Profile Adjustment Algorithm (SPAA) is introduced.
This algorithm combines the base speed profile of the PLDF with the speed profile demonstrated by the driver.
This is done instead of simply using the demonstration by the driver due to multiple reasons.
First, as explained in subsection \ref{subsubsec:limitations}, it is not confirmed that drivers prefer their manual driving style also as their automated driving style.
Instead, a cooperative setting should be used to deduce an appropriate speed profile.
Second, in subsection \ref{subsubsec:intervention_states}, it is explained that humans often show a delayed and overreactive response to system behavior they are not satisfied with \cite{Bi2020,Spencer2022}.

The highlighted algorithm learns based on demonstrated driver interventions.
Therefore, all driver interventions must be classified beforehand and only voluntary interventions within the PLDF's ODD should be used as input for the SPAA.
Reference \cite{Schwager2025_classification} details how such an automatic classification of driver interventions could be applied in a real world scenario.
However, in this work, the classification step is skipped due to the study design described in chapter \ref{sec:study_design}, which does not contain any scenarios outside of the PLDF's ODD.
Therefore, all driver interventions are regarded as relevant voluntary interventions.

\subsubsection{Pedal Intervention Adjustment}

\begin{figure}[t]
    \includegraphics[width=\columnwidth]{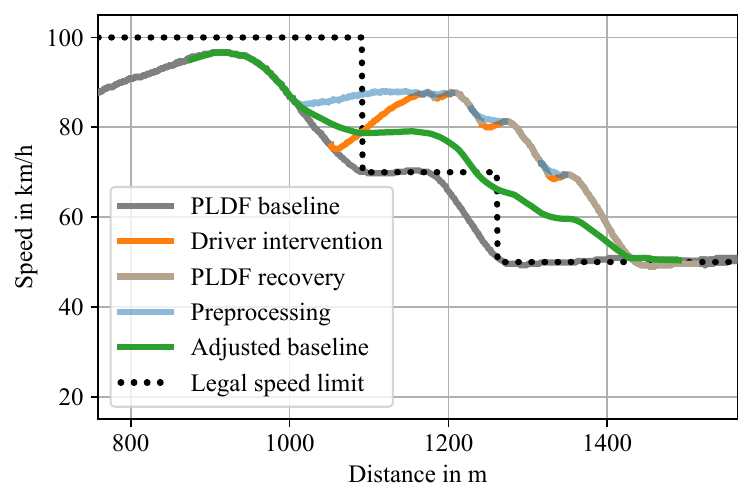}
    \centering
    \caption{Speed-over-distance plot of the SPAA for pedal interventions in a scenario with two successive speed limit decreases.}
    \label{fig:speed_profile_adjustment_algorithm}
\end{figure}

An example for the SPAA for pedal interventions is shown in figure \ref{fig:speed_profile_adjustment_algorithm}.
The underlying driving data was recorded in \cite{Schwager2024_study} and represents naturalistic driving behavior with the original PLDF on real roads.
Two consecutive speed limit decreases are shown with the legal speed as the black dotted line and the PLDF base profile as the gray solid line.
As can be seen, the PLDF's baseline speed always stays at or below the legal speed.
However, the driver seems to be dissatisfied with the PLDF's behavior and intervenes accordingly by pressing the gas four consecutive times, as shown in the orange highlighting.
After each intervention, the PLDF takes back the vehicles control and starts to recover the vehicle speed to its intended baseline speed, as illustrated in the taupe-colored line.
After recording this driver intervention during the study covered in \cite{Schwager2024_study}, the driver annotated it by stating that they would have preferred a later and smoother deceleration in this situation.

Directly replaying this human demonstration would not satisfy this underlying driver preference.
However, it is also not possible to precisely derive the underlying driver preference directly from this intervention.
Instead, it must be approximated.
This approximation is shown in the green solid line in figure \ref{fig:speed_profile_adjustment_algorithm}, which contains the output of the SPAA.
The green speed profile is a mixture of the PLDF's base speed profile and the driver intervention speed profile that takes into account the delayed and overreactive nature of human behavior.

The adjusted speed is calculated as follows:
In the first step of the algorithm, the pedal intervention speed profile is stretched in negative distance direction, i.e., to the left in figure \ref{fig:speed_profile_adjustment_algorithm}.
This is done for each segment where the driving function is inactive, i.e., the orange highlighted segments.
The stretching of the interventions is conducted to compensate the delayed correction behavior of humans, as proposed in \cite{Bi2020,Spencer2022}.
This way, the bad states which cause the driver to intervene are also overwritten by the new speed profile.

Let the original set of distances of a driver intervention be $(d_i)_{i=0}^n$ and the corresponding velocities $(v_i)_{i=0}^n$.
The so-called \textit{stretch factor} $\alpha$ is used for stretching the speed profile.
In this work, a value of $\alpha = 0.5$ is chosen based on qualitative analyses conducted on intervention data from pre-studies.
However, the stretch factor is limited so that the first distance value of the intervention speed profile $d_0$ is moved at most a distance corresponding to 3 seconds.
This limitation is chosen so that a long intervention of multiple seconds does not overwrite too much of the previous speed profile.
Additionally, the stretch factor is reduced around high curvature road segments to ensure a safe learning behavior there.
Using the original distance values $d_i$, the stretch factor $\alpha$, and the maximum distance $d_n$ of the driver intervention, the distance values $(d'_i)_{i=0}^n$ of the stretched profile can be calculated as:

$$
d'_i = d_i - \alpha \cdot (d_n- d_i) \qquad i=0,\dots,n. \eqno{(1)}
$$

In the next step, a linearly decreasing offset is added to the speed values $v_i$ of the intervention speed profile, so that the first speed value $v_0$ at $d'_0$ aligns with the original driver speed profile.
This is done to remove jumps in the profile.
Let the original speed profile driven by the driver be $v_\text{driver}(d)$.
The speed difference in $d'_0$ is then $\Delta v = v_\text{driver}(d'_0) - v_0$.
The speed values with offset are then calculated as follows:

$$
v'_i = v_i + \Delta v \cdot \left (1 - \frac{d'_i - d'_0}{d'_n - d'_0} \right ) \qquad i=0,\dots,n. \eqno{(2)}
$$

The resulting speed profile of this preprocessing is shown in figure \ref{fig:speed_profile_adjustment_algorithm} as the light blue line.
As illustrated, the speed values of the interventions are stretched to the left and they align correctly with the rest of the driver speed profile.
This preprocessed intervention speed profile is then used to calculate the adjusted speed profile shown in green.
Instead of using the individual speed profiles of each pedal intervention, the entire speed profile segment is used where the preprocessed driver speed deviates from the PLDF baseline.
This preprocessed driver speed profile is denoted as $v_\text{prepro}(d)$.
The mean between both profiles is calculated as:

$$
v_\text{mean}(d) = \frac{v_\text{PLDF}(d) + v_\text{prepro}(d)}{2}. \eqno{(3)}
$$

$v_\text{mean}(d)$ is then smoothed using a second-order Savitzky-Golay filter.
The mean between $v_\text{PLDF}(d)$ and $v_\text{prepro}(d)$ is used to compensate the overreactive behavior of human drivers.
This way, a speed profile is generated which is closer to the driver's preferences than the original PLDF speed profile, but is still smoother than the driver's demonstration.
The resulting speed profile is the final adjusted speed, represented by the green line in figure \ref{fig:speed_profile_adjustment_algorithm}.

The algorithm is intended to be used in an iterative manner, with drivers completing multiple drives with the PLDF on the same route.
After each drive, the speed profile is updated based on the driver interventions, and during the next drive the adjusted speed is used as the PLDF's new baseline speed instead.
If the driver is still not satisfied with the behavior, they can intervene again and iteratively train the PLDF until they are satisfied with its speed profile.

\subsubsection{Set Speed Intervention Adjustment}

The above shown mixture of PLDF baseline and driver speed profile is only used for pedal interventions. % since these often show the delayed and overreactive driver behavior.
For set speed interventions, the changes to the speed profile are directly taken over from the driver behavior without merging.
Only if the set speed adjustment was executed shortly after a new legal speed, the set speed offset is set for the whole segment instead. % Potentially remove everything below this line to save space --------------------------------------------

\section{Test Group Study Design} \label{sec:study_design}

To test the applicability of the proposed SPAA, a test group study is conducted in a simulation environment.
In the study, the base PLDF and the adaptive PLDF using the SPAA are directly compared.
The goal of the study is to test two hypotheses:

\begin{enumerate}
    \item The driver satisfaction with the PLDF is increased by the proposed system compared to the baseline system.
    \item The frequency of driver interventions is reduced by the proposed system compared to the baseline system.
\end{enumerate}

\subsection{Participants}

The main test group study was conducted with 43 valid participants.
The participants were recruited internally at ANONYMIZED from various departments.
A requirement for study participation was that subjects identified as regular users of driver assistance systems.
This requirement was chosen so that participants would not first have to learn how to use an ADAS during the study. % but already had experience with similar systems in their private life.
The age and gender distribution of the participants is shown in table \ref{tab:gender_age_distr}.

\begin{table}[b]
    \renewcommand{\arraystretch}{1.2}
    \caption{Gender and age distribution of the participants.}
    \label{tab:gender_age_distr}
    \begin{center}
        \begin{tabular}[width=\columnwidth]{ccccc}
            \toprule 
            \centering  Gender &   Age\,$\le$\,30   &   30\,$<$\,Age\,$\le$\,40    &   40\,$<$\,Age\,$\le$\,50    &   Age\,$>$\,50\tabularnewline
            \midrule
            \midrule
            \centering  Female          &   8                       &   4                               &   0                               &   2\tabularnewline
            \midrule
            \centering  Male            &   9                       &   7                               &   6                               &   7\tabularnewline
            \bottomrule
        \end{tabular}
    \end{center}
\end{table}

\subsection{Simulation Environment}

\begin{figure}[b]
    \includegraphics[width=\columnwidth]{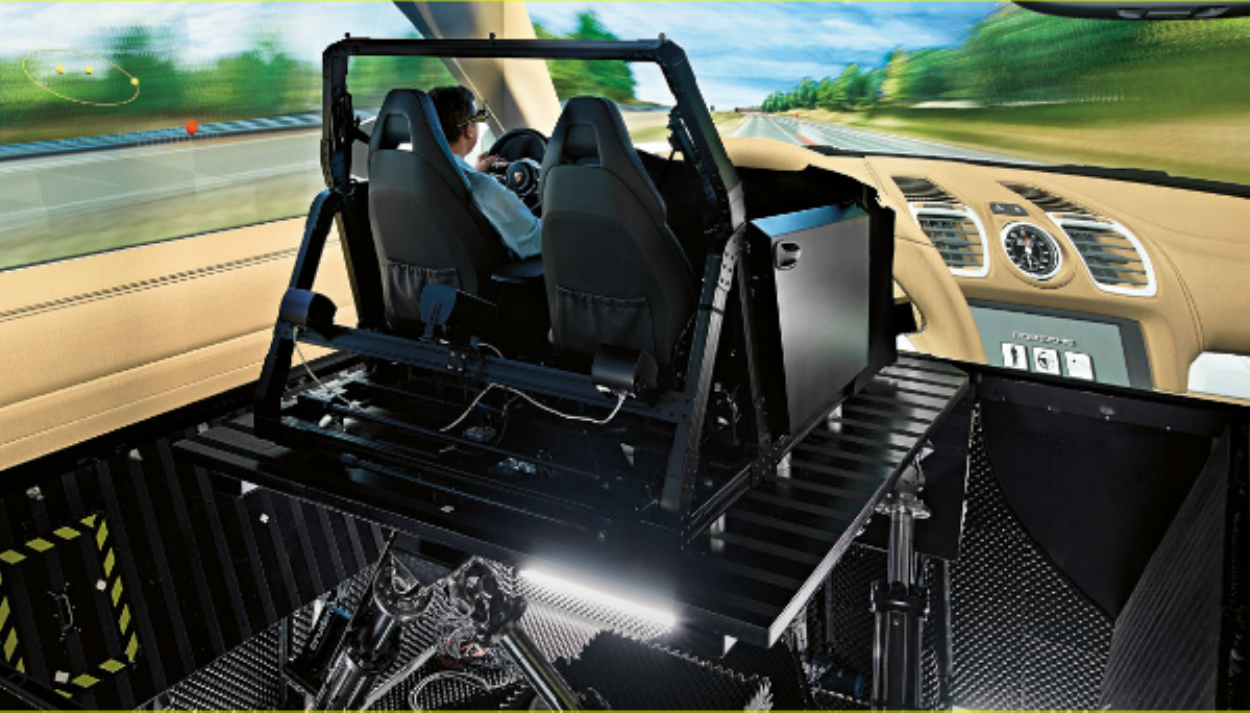}
    \centering
    \caption{The simulator used in this study.}
    \label{fig:simulator}
\end{figure}

The simulator used for this study is shown in figure \ref{fig:simulator}.
As can be seen, it is composed of a hexapod with a movable platform.
A vehicle cockpit is mounted on top of the platform and high resolution LED walls are mounted around it.
With the hexapod, longitudinal and lateral accelerations as well as rotations are simulated.
Depending on the driven speed in the simulation, road noises and vibrations are simulated as well to further increase realism.

\begin{figure}[b]
    \includegraphics[width=\columnwidth]{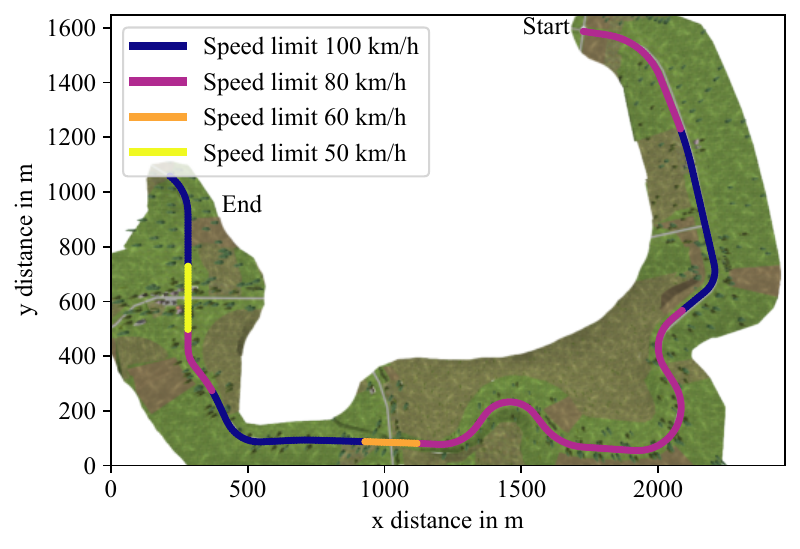}
    \centering
    \caption{Simulation map used for the study with highlighted speed limits.}
    \label{fig:xy_map}
\end{figure}

The used simulation map is a curvy rural road shown in figure \ref{fig:xy_map}.
The legal speed limits on the track are highlighted in the plot, consisting primarily of \unit[80]{km/h} and \unit[100]{km/h}.
The map also contains a short bridge over a highway and a small village with speed limits of \unit[60]{km/h} and \unit[50]{km/h} respectively.
Driving through this \unit[4.5]{km} long track takes approximately 3 minutes and 30 seconds.

\begin{figure}[t]
    \includegraphics[width=\columnwidth]{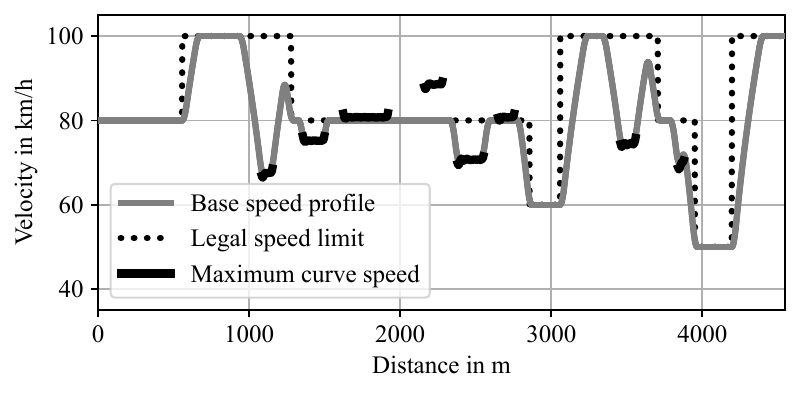}
    \centering
    \caption{Velocity over distance plot of the the speed profile of the base PLDF without interventions on the simulation track.}
    \label{fig:vs_base}
\end{figure}

For the test group study, a mockup of the PLDF was created which behaves equivalently to its in-production version.
The user interface of the PLDF is also the same as in the real vehicle and features a control lever which can be used to activate and deactivate the PLDF.
Set speed offsets can also be added using said lever.
When driving on the simulation track with the activated base PLDF, the speed profile shown in figure \ref{fig:vs_base} is driven.
As can be seen, the PLDF always stays at or below the legal speed.
The curvatures from the map are also taken into account by calculating maximum curve speeds.
These curve speed constraints are highlighted as black bars.

\subsection{Study Procedure}

Before driving in the simulator, participants were briefed about the study.
They were told that the focus lies on the evaluation of two versions of the same PLDF, system A and system B.
System A is the base PLDF, while system B represents the adaptive version.
Drivers were not informed about the difference between the two systems.
However, in order to minimize confusion during the study, they were told that the driving functions adjust themselves to the drivers' preferences in some way.
The briefing also includes safety instructions for the simulator and how to use the PLDF.
The drivers then carried out two test drives to familiarize themselves with the simulator, the driven route, and the PLDF before the actual experiment.

After the introduction, drivers started with the system A tests.
They were instructed to drive with the PLDF as they would in real life and that they are free to adjust its behavior based on their preferences with the pedals and set speed changes.
During these drives, they used the base PLDF which does not use the SPAA.
After two drives with interventions, they were told to drive one last time without interventions.
As the base system does not adapt to the driver's behavior, the base PLDF speed profile is driven again. %drives its base profile shown in figure \ref{fig:vs_base}.
Afterward, the drivers filled out two questionnaires about their satisfaction with the system.
The questionnaires are covered in chapter \ref{subsec:questionnaire}.

After the system A tests, drivers continued with the tests of system B, which is the adaptive PLDF.
Using the interventions recorded during the second drive with system A, the adaptive PLDF was pre-trained.
The system B tests consisted again of two drives with interventions and one drive without interventions to experience the final speed profile.
However, this time the PLDF learned from the driver interventions and updated its behavior after every drive.
After the test drives, the drivers filled out the same questionnaires for system B.

Finally, drivers were asked for free form feedback on the system which was documented as full text.
This feedback was used in a qualitative analysis of the study results, covered in \ref{subsec:qual_analysis_limitations}.
In total, the study took one hour for each participant.

A limitation of this study design is that the experiment order had to be fixed to first A then B, because system B is trained on the interventions from the system A tests.
A randomization of the experiment order would only have been possible by significantly increasing the study duration.
However, the duration was consciously limited to one hour to ensure participant concentration and to reduce the likelihood of motion sickness.
While this study design is prone to learning effects, it realistically resembles the order in which customers would experience the system.

\subsection{Questionnaire} \label{subsec:questionnaire}

Two questionnaires were used for the evaluation of each system.
First, the satisfaction subscale of the system acceptance questionnaire introduced in \cite{Laan1997} is used.
Second, a custom questionnaire is used to evaluate specifically the speed profile driven by both functions.
There, for all questions a five point Likert scale is used ranging from strongly agree (1) to strongly disagree (5).
The questionnaire encompasses the following six statements which inquire about the satisfaction with specific aspects of the speed profile:
"I was satisfied with the system's..."

\begin{itemize}
    \item Speed on straight road segments.
    \item Acceleration timings onto higher speed limit signs.
    \item Deceleration timings onto lower speed limit signs.
    \item Speed in and around curves.
    \item Acceleration strength.
    \item Deceleration strength.
\end{itemize}

These statements are derived from the common intervention types with the PLDF found in \cite{Schwager2024_study,Schwager2024_needed_adjustments}.

\section{Results} \label{sec:results}

In this chapter, the results of the test group study are explained.
The main focus lies on the testing of the two hypotheses introduced in chapter \ref{sec:study_design}.

\subsection{Driver Satisfaction Analysis}

Using the questionnaires introduced in chapter \ref{subsec:questionnaire}, the satisfaction of the drivers with both the base PLDF and the adaptive version are measured.
The first satisfaction measure is the \textit{general satisfaction} with the driving function, which corresponds to the satisfaction subscale of the system acceptance questionnaire introduced in \cite{Laan1997}.
By calculating the mean value of the respective four items, the general satisfaction score is calculated for both systems for each driver.
Directly comparing the general satisfaction scores yields the result that 34 participants gave system B a better rating than system A, while three participants gave the exact same score for both systems, and six participants gave system B a worse rating than system A.
The reasons for these worse ratings are examined further in chapter \ref{subsec:qual_analysis_limitations}.
When calculating the mean ratings for system A and B over all participants, system A reaches a score of 3.69 while system B reaches 4.27, where 5.0 is the best possible score and 1.0 is the worst score.
This results in an average score improvement of 0.58.
Using a Student's t-test for paired samples, a significant difference between the satisfaction ratings of system A and system B is found ($t(42)=4.60$, $p=3.85 \cdot 10^{-5}$) %($W=103$, $p=2.18 \cdot 10^{-5}$).

The second satisfaction measure used in this study is the \textit{speed profile satisfaction}, which was measured using the custom questionnaire introduced in chapter \ref{subsec:questionnaire}.
The speed profile satisfaction for each driver is again calculated using the mean scores of the questionnaire ratings, ranging from 5.0 (best) to 1.0 (worst).
In total, 40 participants rated the speed profile of system B higher than system A with only 3 participants preferring system A.
The mean satisfaction scores are 3.06 for system A and 4.14 for system B, which is an average improvement of 1.08 using system B.
A Wilcoxon signed-rank test confirms that this difference is significant ($W=45.0$, $p=3.83 \cdot 10^{-9}$).

With these results, the first hypothesis can be confirmed:
The adaptive PLDF using the proposed SPAA increases the driver satisfaction significantly compared to the base PLDF.

\subsection{Intervention Rate Analysis}

The second hypothesis states that the frequency of driver interventions should be reduced with system B compared to system A.
To evaluate this, Intervention Rates (IRs) are calculated for the pedals and the set speed.
An IR is defined as the relative time of each drive during which the system was intervened in.
The pedal IR is therefore calculated by summing up the duration of all pedal interventions over one drive and dividing it by the total driven time.
Similarly, the set speed IR is calculated as the relative time where a set speed offset is active.
In the context of system B, which takes over previous set speed inputs automatically in the next iteration, the set speed IR is calculated using only new set speed interventions from the current iteration.
Additionally, a combined IR is calculated as the relative time any type of intervention is active.

\begin{figure}[t]
    \includegraphics[width=\columnwidth]{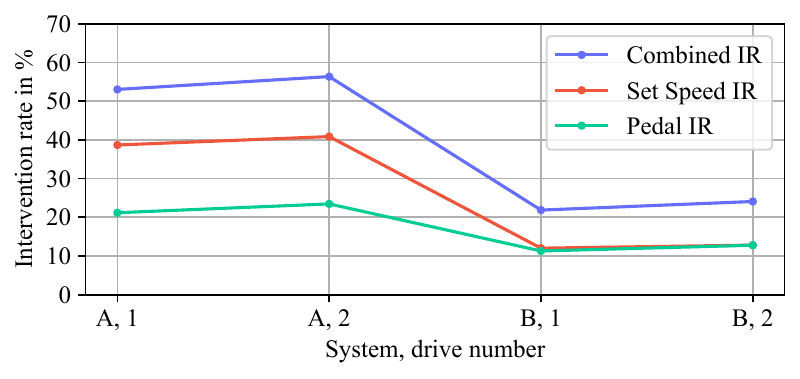}
    \centering
    \caption{Intervention rate evolution over multiple drives.}
    \label{fig:intervention_rate_time_based_evolution}
\end{figure}

With this logic, the IR of each drive can be calculated and analyzed.
For each participant, two drives with interventions were recorded per system, resulting in a total of four drives.
The development of the pedal IR, set speed IR, and combined IR over the four drives is depicted in figure \ref{fig:intervention_rate_time_based_evolution}.
As illustrated, the average combined IR for system A at \unit[54.68]{\%} is much higher than for system B with an average of \unit[22.97]{\%} combined IR.
A Wilcoxon signed-rank test proves that this reduction is significant ($W=1$, $p=4.55 \cdot 10^{-13}$).
The pedal IR drops from \unit[22.32]{\%} for system A to \unit[12.04]{\%} for system B, which is significant according to a Student's t-test ($t(42)=6.42$, $p=9.78 \cdot 10^{-8}$).
Interestingly, the strongest effect can be seen in the set speed IR, which drops from \unit[39.76]{\%} to \unit[12.42]{\%}, which is significant according to a Wilcoxon signed-rank test ($W=24$, $p=1.38 \cdot 10^{-7}$).
This means that both pedal interventions and set speed interventions are significantly reduced by the SPAA.
As it can be seen in figure \ref{fig:intervention_rate_time_based_evolution}, already one iteration of the SPAA can significantly reduce the IR.

While the IRs are significantly reduced by the proposed system, they do not reach \unit[0]{\%} during the test group study.
Accordingly, driver still intervene while using system B after two iterations of SPAA applications.
Thus, the long-term driving behavior with system B should be investigated in future work to evaluate, whether the IRs ever converge to \unit[0]{\%}.
Due to the limited study duration of one hour, this could not be investigated in the present study.

\subsection{Example Profile Development}

\begin{figure}[t]
    \includegraphics[width=\columnwidth]{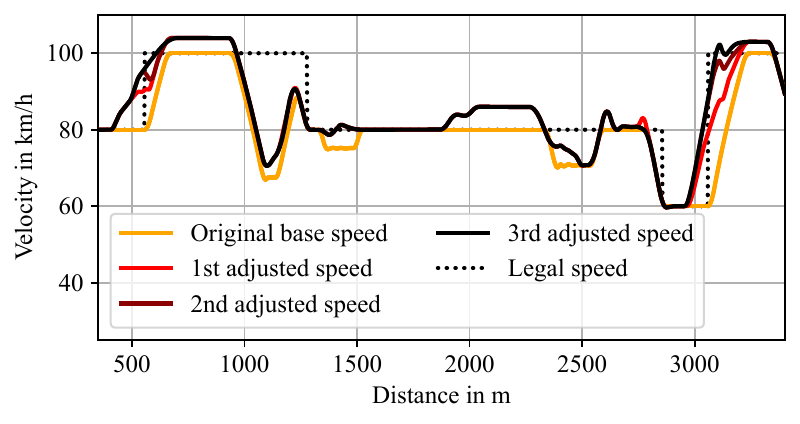}
    \centering
    \caption{Velocity over distance plot of the development of the speed profile for one example driver.}
    \label{fig:profile_development_plot}
\end{figure}

In figure \ref{fig:profile_development_plot}, the development of the speed profile of one example driver is shown.
During the speed limit increases at \unit[500]{m} and \unit[3000]{m}, it can be seen how the speed profile is adjusted every iteration due to repeated driver interventions.
In these segments, multiple interventions were needed to adjust the speed profile better to the driver's wishes.
However, in all other segments, the driver only intervened during the first drive, indicating that they were already satisfied with most of the adjustments made by the SPAA after running it once.
This is a good example for the significantly reduced IR in the drives with system B compared to system A.

\subsection{Qualitative Feedback Analysis} \label{subsec:qual_analysis_limitations}

The free form feedback received from each participant was used to analyze the strengths and weaknesses of the system and to find further improvement potential.
Most of the received feedback was generally positive and many drivers realized how the system works while driving.
The driving style of system B was generally described as more natural and human-like compared to the more artificial and machine-like driving style of system A.
However the most common point of criticism was the lack of communication with the driver.
Many participants mentioned that they would have preferred to receive more information about the decisions of system B.
For example, when showing a legal speed and a set speed of \unit[70]{km/h}, the system could still drive above or below this speed based on previous driver interventions.
In such cases, drivers would have preferred, e.g., an icon or a small text in the instrument cluster to explain this behavior.
Also, the fact that the system learns from interventions should be stated explicitly to real customers to reduce confusion.
Otherwise, drivers may feel unsafe if the PLDF behaves in a way they do not understand.
Multiple drivers stated this lack of communication was their primary reason for assigning a higher general satisfaction to system A despite rating system B's speed profile as more desirable.

Due to the design of this study, every driver intervention was directly learned by system B without arbitration, including unintentional driver interventions.
Affected drivers described it as difficult and irritating to overwrite incorrectly learned behavior again.
Moreover, real-life driving with a PLDF may also contain many driver interventions which should not be learned, e.g., because they fall outside of the PLDF's ODD \cite{Schwager2024_study}.
Therefore, an arbitration of the driver interventions is necessary that automatically classifies the reason behind each driver intervention, as introduced in \cite{Schwager2025_classification}, and then accumulates the data over multiple drives, as discussed in \cite{Schwager2024_needed_adjustments}.
Based on these approaches, hotspots of consistent interventions could be derived, where the SPAA is then applied to ensure that only consistent driver behavior is learned.

\section{Conclusion and Limitations} \label{sec:summary_and_outlook}

In this paper, an algorithm for the location-based adjustment of PLDF speed profiles based on driver interventions is proposed.
The SPAA takes into account the PLDF's base speed profile and the driver demonstrations by combining both into a new speed profile.
This interactive approach can be applied iteratively until the individual drivers stop intervening and a fitting speed profile is found.
The applicability of the system is shown in a test group study where a significant increase in driver satisfaction and a significant decrease in the intervention frequency of the participants was found.
The personalization approach of learning from feedback in the form of driver interventions was therefore confirmed to be effective.

However, due to the limited study duration, the driver behavior with the adaptive PLDF over longer durations could not be evaluated.
Future work should therefore address this long-term driving behavior and investigate whether the SPAA is applicable to completely reduce the IR of drivers to \unit[0]{\%} or whether further adjustments are necessary.
Another limitation of the employed study design is that the order of the experiments had to be fixed as A then B.
As this could lead to learning effects in the participants, future work should randomize the experiment order.
Future work should also focus on the integration of the proposed SPAA in a fully working prototype in a real vehicle.
As described in chapter \ref{subsec:qual_analysis_limitations}, this would include the automatic classification of driver interventions and their accumulation over multiple drives.
By using these arbitration steps, it can be ensured that only relevant and consistent interventions are learned.
A study conducted in a real vehicle would also address potential simulation-to-reality gaps.

In the feedback discussions following the experiment, drivers most commonly criticized a lack of communication in system B. % about the system behavior and reasoning.
To reduce confusion with the system, a suitable communication interface with the driver, e.g., in the instrument cluster must also be designed.

\bibliography{IEEEabrv, driving_function_prototype_references}

\end{document}